# An Experiment with the Use of ChatGPT for LCSH Subject Assignment on Electronic Theses and Dissertations

Eric H. C. Chow[a]*, TJ Kao[b] and Xiaoli Li[b]

[a]Hong Kong Baptist University, Hong Kong, China; [b]University of California Davis, U.S.A.

Address correspondence to Eric H. C. Chow, Digital Scholarship Manager, Au Shue Hung Memorial Library, Hong Kong Baptist University, 34 Renfrew Road, Kowloon Tong, Kowloon, Hong Kong. Email: choweric@hkbu.edu.hk*

**Abstract**

This study delves into the potential use of large language models (LLMs) for generating Library of Congress Subject Headings (LCSH). The authors employed ChatGPT to generate subject headings for electronic theses and dissertations (ETDs) based on their titles and abstracts. The results suggests that LLMs such as ChatGPT have the potential to reduce cataloging time needed for assigning LCSH subject terms for ETDs as well as to improve the discovery of this type of resource in academic libraries. Nonetheless, human catalogers remain essential for verifying and enhancing the validity, exhaustivity, and specificity of LCSH generated by LLMs.

**Introduction**

*MARC and LCSH Systems*

Library metadata practitioners usually produce bibliographic records for library resources in the form of MARC (Machine-Readable Cataloging) records. MARC is a data format that organizes information about books and other materials in a library collection. One important part of the data is the MARC 6XX Subject Access Fields, which are used to input subject access entries and terms (e.g., topical terms, personal names, places, and time periods) covered by each book or resource. Most of these fields contain subject terms based on controlled vocabularies, one of which is the Library of Congress Subject Headings (LCSH).

A key feature of the LCSH system is the application of subdivisions, allowing catalogers and indexers to construct very precise and multi-faceted subject strings by combining main headings with relevant topical, form, chronological and geographic subdivisions in a systematic way. [1,2] The rules for heading construction presuppose that catalogers have a comprehensive understanding of the LCSH structure and principles. Yet, it takes a considerable amount of time and effort to train new catalogers to correctly apply these elaborate and intricate rules. Accordingly, catalogers would undergo years of training before they can independently create complex strings; in many cases, training is not provided at all, leaving catalogers to rely on existing bibliographic records as examples.

This complexity is not just a barrier for the enrolment of new catalogers, but also an ongoing challenge for the already trained professionals. With the evolution of the system and incorporation of new terms, constructing accurate and relevant subject strings demands a yet higher level of expertise and continuous learning. In addition, based on our understanding of the Library of Congress' *Subject Heading Manual (SHM)* and experience in computer programming, the intricate nature of the LCSH system, which requires significant human intervention and judgment, makes it difficult to automate the system using rule-based computer programs. This reliance on manual labor not only increases the time and resources needed for cataloging, but also increases the risk of inconsistencies and errors entering the system, especially in large and diverse collections.

*Automatic Cataloging Record Generation*

To date, seeking to alleviate the effort of human catalogers, several previous studies have attempted using machine learning in automatic assignment of subject headings on library collections. For instance, Wartena et al.[3] focused on creating classifiers for frequently used terms in abstracts and extracting headings directly from texts. However, despite achieving acceptable levels of precision, the classifiers had limited recall due to the inability to train on a vast number of LCSH terms. Drawing conclusions based on these findings, the authors highlighted the challenges in categorizing documents with the LCSH due to the vast number of headings and specificity of many terms. In another relevant study, Kazi et al.[4] explored the LCSH assignment task by formulating it as a multi-label classification problem and developing predictive models using a gold-standard dataset. The results revealed the viability of predicting the LCSH for scholarly articles using models like Decision Trees and Artificial Neural Networks.

In recent years, the surge in the development of large language models (LLMs) has fueled their use in various machine learning applications, including label assignment on a given text input. LLMs such as OpenAI's ChatGPT are AI tools built upon multi-layer recurrent neural networks trained on vast amounts of text data[5] to generate human-like text output. LLMs use transformer-based architectures and statistical methods to predict subsequent words given a textual input.[6] LLMs also display remarkable proficiency in composing computer programming code, simulating scientific discourse, and even creating literary works such as prose and poetry. Similarly remarkable is LLMs' current performance on labeling and classification tasks on a given textual input, such as sentiment analysis,[7] political social media posts classification,[8] and topic classification.[9] At the time of this writing, both OpenAI and Google have released multimodal versions of their LLMs, which can take image, video, and text as inputs to produce generative text output.

Recently, LLMs have also come to be used in the area of library cataloging, which marked a significant shift from previous methods by eliminating the need for machine learning model training. In one of these recent studies, Brzustowicz[10] tested ChatGPT's ability to generate MARC records. The experiment involved using ChatGPT to create MARC records for a variety of items, including books and sound recordings, in different languages. In order to test ChatGPT's ability to generate an original record, the selected items included both those with existing MARC records in OCLC's WorldCat and those without a corresponding entry. The results revealed that ChatGPT successfully produced MARC records closely resembling those created by professional human catalogers. The findings also demonstrated ChatGPT's proficiency in extracting essential metadata, such as title and authors. However, discrepancies were noted in the subject headings assignment. An interesting finding in Brzustowicz's study was that ChatGPT managed to generate original MARC records when no existing WorldCat entry was available, showing its potential for creating new catalog entries.

Furthermore, Zhang et al.[11] specifically tackled the problem of subject heading assignment using ChatGPT. The authors designed a prompt that included a list of 22 subject heading (such as "chemical science," "engineering," "law and legal studies") from the Australian and New Zealand Standard Research Classification (ANZSRC) scheme, along with instructions on the use of this scheme and several demonstration examples (known as few-shot prompting technique). The results showed that, while ChatGPT demonstrated high precision in certain subject disciplines, lower performance was observed in others. This variation in performance highlights the challenges in assigning labels to multidisciplinary research datasets, as well as the limitations of in-context learning methods in applying discipline-specific classification rules.

Yet, a limitation of Zhang et al.'s study[12] is that may not have capitalized the internal corpus of a LLM – specifically, MARC records and the LCSH that may have already been internalized in an LLM – for library cataloging functions. In fact, LLMs such as ChatGPT are trained on text data publicly available from the Internet.[13] The LCSH and MARC records are readily available on the Internet's public domain. At the time of this writing, a total of 459,624 records of authorized LCSH topical vocabulary are freely and openly accessible on the Library of Congress website.[14] For instance, the LCSH *Hong Kong Sign Language* and related information such as variant terms, broader terms, and associated book titles are made readily available on the Library of Congress website.[1] Moreover, many academic libraries and commercial publishers have made the MARC 21 records of their collections open-access.

These records frequently contain detailed bibliographic metadata, including titles and the associated LCSH subject headings, most of which with subdivision strings. Therefore, it is highly plausible that LLMs such as ChatGPT have already ingested from the public domain a vast amount of metadata encoding the relationships between thousands of LCSH vocabularies and bibliographic elements from millions of existing MARC 21 records.

---

[1] See https://id.loc.gov/authorities/subjects/sh2007020076.html

*Subject Analysis for Theses and Dissertations*

Theses and dissertations are original research outputs that not only reflect the quality of students' and faculty members' scholarly works, but also represent a body of knowledge and history of the corresponding degree-granting institution.[15]

In this context, it is important for research institutions to properly catalog theses and dissertations, making them more discoverable. Unfortunately, such a task is frequently highly time-consuming even for experienced catalogers, as it involves a highly complex subject analysis of writings on sometimes very narrow and specialized subject areas. To reduce processing time, it is not uncommon for academic library catalogers to input author-supplied keywords into the 653 fields for uncontrolled index terms.[16] Yet, these author-supplied keywords are often unhelpful for the LCSH assignment,[17] as students are not professionally trained on subject analysis, and the keywords are most likely non-conformant to the LCSH controlled vocabularies that can be used directly in the 600, 610, 630, 650, and 651 fields in a MARC record. Furthermore, although most theses and dissertations nowadays are born-digital (Electronic Theses and Dissertations, or ETDs), in which case many fields of the MARC record can be populated automatically, subject heading assignments in 6XX fields, especially using LCSH controlled vocabularies, still remain untouched, which results in a demanding task for catalogers. Aiming to address this concern, in the present study, we specifically examine the use of ChatGPT for subject heading assignment for theses and dissertations, as the prospective results would be highly informative and beneficial not only for catalogers of research institutions, but also for entire research communities.

**Methodology**

In this study, we attempted to use ChatGPT's internalized corpus of the LCSH, supplemented by three examples to describe the output in MARC 21 format. To address the shortcoming discovered in Brzustowicz's[18] work, we focus specifically on LC Subject Headings assignment. Furthermore, unlike Zhang et al.,[19] we did not include explicit heuristics and list of subject headings in the prompt. The prompt used in this study was as follows:

> As a library cataloger, please provide the appropriate Library of Congress (LC) subject heading(s) that best match the abstract and title of the given work. The detail of the output is as follows:
> - Format the subject headings in MARC 21 (fields 650, 600, 610, 651, 630) without using field 880.
> - Assign subject headings that best summarizes the overall contents of the work, and as specific as possible on the topics being covered.
> - Include multiple subject headings if the book covers various topics.
> - Limit the number of subject headings between one to four.
> - Provide the MARC fields only, without explanations or notes or greetings.
>
> --- Start of Examples ---
>
> Example 1:
> Title: Composite property rights and boundary-treading resistance : a case study of County in Eastern Sichuan
> Summary: This thesis studies land expropriation disputes from the angle of property right, exploring its origins from the relationships between township (town) government and villagers, village collective and villagers, and different villagers, focusing on peasants' resisting low land expropriation.
>
> 650  $aLand tenure $zChina.
> 650  $aRight of property $zChina.
>
> Example 2:
> Title: Determinants of expertise of Olympic style Taekwondo performance

Summary: The purpose of this study was to identify the determinants of expertise and the contributory effect of domains to the Olympic style Taekwondo performance. Eighty-seven Taekwondo athletes with different levels of expertise, namely elite, sub-elite and practitioner were recruited.

650 $aAthletes $xPsychological aspects.

650 $aAthletes $xResearch.

650 $aTae kwon do.

Example 3:

Title: An ADMM approach to the numerical solution of state constrained optimal control problems for systems modeled by linear parabolic equations

Summary: We address in this thesis the numerical solution of state constrained optimal control problems for systems modeled by linear parabolic equations. For the unconstrained or control-constrained optimal control problem, the first order optimality condition can be obtained in a general way.

650 $aFinite differences.

650 $aMultipliers (Mathematical analysis)

650 $aNumerical analysis.

--- End of Examples ---

With the growing popularity of LLMs, researchers and practitioners alike have experimented and discovered general rules and best practices in crafting LLM prompts that would produce meaningful responses. Accordingly, in crafting our prompt, we took into account the following considerations. A good prompt should contain four main components: context, instruction, output indicator and input data.[20] Therefore, our prompt design captured the first three elements – context ("*As a library cataloger…*"), instruction ("*…provide the appropriate Library of Congress Subject Headings…*"), and output indicator (the examples demonstrating how the output should look like). The input data, which consisted of the title and abstract of the ETD in question, was then appended to context, instruction, and output indicators. We also considered available evidence showing that, by repetitively providing several examples of the desired output, LLMs could produce better and more accurate output.[21] Accordingly, our prompt included three examples of the desired output in the MARC format. Furthermore, an effective prompt should be specific and include constraints whenever necessary (e.g., "*Provide the MARC fields only, without explanations or notes or greetings*"). During the experimentation in crafting the prompt, non-Latin input data (e.g., Chinese title and summary) were used, and the prompt resulted in the creation of the 880 field, which was used as linkage to the 6XX field for subjects in non-Latin scripts. As the ETD records examined in our experiment were in English (Latin scripts) only, the specific instruction of "*...without using the 880 field*" was included in the prompt.

The experiment was conducted on January 17, 2024 using ChatGPT-3.5 through Microsoft Azure OpenAI API. In order to reduce randomness in the responses, the LLM "temperature" setting was set to 0 (zero). The MARC records of 30 ETDs were randomly selected from the WorldCat and UC Davis Library catalog. The ETDs covered various disciplines and subject matters. A Python programming script[2] was written to feed the prompt presented above, appended sequentially by the title and abstract of each ETD record into ChatGPT. The total processing time for the 30 ETD records amounted to approximately 3 minutes.

**Results**

ChatGPT produced between three to five subject headings (mostly with subdivisions) for each of the ETD records. The results were tabulated into an Excel spreadsheet[3] for a subsequent analysis by an experienced professional cataloger with respect to the following three principles described in *Basic Subject Cataloging Using LCSH – Instructor's Manual* published by the Library of Congress:[22]

---

[2] See https://github.com/choweric/llm-lcsh-experiment/

[3] See https://github.com/choweric/llm-lcsh-experiment/raw/main/results.xlsx

1. Is the MARC coding correct?
2. Is the heading string valid? (e.g., are the elements authorized by authority records or by the Subject Heading Manual? Is the order and placement of subdivisions appropriate?)
3. Do the headings assigned reflect the content of the work being cataloged?

*Correctness of MARC Coding*

Among the 30 analyzed ETD records, ChatGPT's outputs of MARC coding were mostly correct (90%), with only three records containing incorrect codes. Specifically, these three erroneous records contained subject headings that were assigned to 650 fields where it was supposed to be 600 or 610.

*Validity of LCSH Assignment*

ChatGPT produced valid LCSH for seven ETD records, or 23.3%, that completely conforms to the LCSH controlled vocabularies; in the remaining 23 ETD records, the assigned LCSH had varying degrees of validity. In what follows, we provide a qualitative summary of the results.

   a.  Topical terms that were clearly not from the LCSH controlled vocabulary: *Abuse, Adaptations, Environmental regulation, Ethnohydrology, Freezing resistance of plants, Growth mindset, Infidelity in motion pictures, Laser annealing, Princess Nourah bint Abdulrahman University, Socioeconomic factors, Women in sports administration.*

   b.  Topical terms that closely matched the existing LCSH controlled vocabulary. For example:

| ChatGPT Output | LCSH Controlled Vocabulary |
|---|---|
| *College personnel* | *College personnel management* |
| *Doping (Semiconductor technology)* | *Semiconductor doping* |
| *High-frequency ventilation* | *High-frequency ventilation (Therapy)* |
| *Urea cycle* | *Urea* |

c.  Subject headings from the LCSH controlled vocabulary, but in their variant form, rather than the preferred terms. For example:

| ChatGPT Output | LCSH Controlled Vocabulary (Preferred Terms) |
|---|---|
| *Animal translocation* | *Animal introduction* |
| *Personal growth* | *Maturation (Psychology)* |
| *Psychoanalytic criticism* | *Psychoanalysis and literature* |
| *Zooarchaeology* | *Animal remains (Archaeology)* |

d.  Topical terms that were not LCSH but provided a more precise or nuanced description than those available in the LCSH controlled vocabulary for the specific topic of the ETD being examined. For example:

| ChatGPT Output | LCSH Controlled Vocabulary |
|---|---|
| *Group counseling for men* | *Group counseling for teenagers* |
| | *Group counseling for children* |
| | *Group counseling for girls* |
| | *Group counseling for teenage girls* |
| *Respiratory distress syndrome, Newborn* | *Respiratory distress syndrome* |
| | *Respiratory distress syndrome, Adult* |

e.     Subdivision terms that did not conform to the LCSH subdivision rules.

For example:

| **ChatGPT Output** | **Reason for non-LCSH conformant** |
|---|---|
| *$a Press coverage $x Political aspects.* | *Press coverage* is a subdivision and cannot be used as a main topical subject. |
| *$a Cardiovascular system $x Risk factors.* | *Risk factors* cannot be used as subdivision of *Cardiovascular systems*. It can only be appended to individual diseases and types of diseases. |
| *$a English language $x Lexicography $z Great Britain $x History $y 18th century.* | *Lexicography* cannot be geographical subdivided |
| *$a Women composers $z France $y 20th century* <br> *$a Women musicians $z France $y 20th century* | Chronological subdivisions, such as *$y 20th century*, can only be appended to topical subjects of historical nature or form / genre headings. |
| *$a Middle Ages $x Literary collections* | *Literary collections* is a form subdivision, not a topical subdivision |
| *... $x Career development* <br> *... $x Factors* <br> *... $x Men* <br> *... $x Sexual orientation* | Invalid subdivision terms |

*Specificity and Exhaustivity of LCSH Assignment*

To find out whether the headings assigned by ChatGPT reflected the content of the work being cataloged, we focused on two key principles of cataloging specified in the Library of Congress' *Subject Heading Manual (SHM) H 180 – Assigning and Constructing Subject Headings*: specificity and exhaustivity. The principle of specificity guides catalogers to assign subject headings that are as precise as the content they represent, while the principle of exhaustivity directs them to assign subject headings that encapsulate the overall content of the work and highlight its significant topics. Accordingly, we initiated the process by scrutinizing titles and abstracts, and subject headings generated by ChatGPT for the 30 MARC records. Out of all the ETDs records being reviewed, 16 (53.3%) were evaluated as acceptable in both specificity and exhaustivity. Note that this evaluation was performed independently from validity evaluation (the strict conformance of the LCSH vocabularies and subdivision rules). In what follows, for illustrative purposes, we provide an analysis of five cases covering a spectrum of specificity and exhaustivity.

**Case 1**
ETD title: Soft-release may not enhance translocations of wild-caught eastern massasaugas (sistrurus catenatus)
ChatGPT output:
    *Wildlife reintroduction – Michigan*
    *Rattlesnakes – Michigan*
    *Animal translocation – Michigan*
    *Animal radio tracking – Michigan*
Analysis: ChatGPT is able to assign subject headings capturing major concepts of the resource optimally both in terms of specificity and exhaustivity.

**Case 2**

ETD title: Career experiences and intentions of women in senior level intercollegiate athletic administration

ChatGPT output:

>  *Women in sports administration*
>
>  *College sports – United States – Management*
>
>  *Career development – United States*
>
>  *Sex discrimination in sports – United States*

Analysis: While *Women in sports administration* is not a valid LCSH, the other headings are valid and properly provide coverage of major subject matters in this resource.

**Case 3**

ETD title: "Mind how you vote, boys" the crisis of economic voter intimidation in the late-nineteenth century United States, 1873-1895

ChatGPT output:

>  *Voter intimidation – United States – History – 19th century*
>
>  *Elections – United States – History – 19th century*
>
>  *Industrialization – United States – History – 19th century*
>
>  *Ballot – United States – History – 19th century*

Analysis: The subject headings assigned by ChatGPT capture the main concept, *voter intimidation*, along with *elections* and *ballot*. However, the inclusion of *Industrialization* is debatable.

**Case 4**

Teachers of the public, advisors to the sultan preachers and the rise of a political public sphere in early modern Istanbul (1600-1675)

ChatGPT output:

>  *Preaching – Turkey – Istanbul – History*
>
>  *Public opinion – Turkey – Istanbul – History*
>
>  *Religious institutions – Turkey – Istanbul – History*
>
>  *Istanbul (Turkey) – History – 17th century*

Analysis: While ChatGPT captures the main concepts, e.g. *preaching* and *Istanbul (Turkey)*, and covers additional ones, e.g. *public opinions* and *religious institutions*, one major theme, political aspect, is missing. This output does not pass the exhaustivity test.

**Case 5**

ETD title: Push-pull Hezbollah the New York Times and the Washington Post news coverage of three Israel-Lebanon conflicts

ChatGPT output:

> *Arab-Israeli conflict*
>
> *Press coverage – Political aspects*
>
> *Lebanon – Politics and government*

Analysis: While ChatGPT is able to identify some major subject matters, e.g. *press coverage*, *politics and government*, it fails to provide headings with appropriate level of specificity.

**Discussion**

A notable outcome of the present study is the high level of correctness ChatGPT demonstrated in MARC field coding, indicating that catalogers can generally depend on ChatGPT for this task, albeit with the understanding that occasional errors may require manual correction through spot checks. Catalogers can also rely on Integrated Library Systems (ILS) or other authority control services to routinely check and identify records with field coding errors, flagging those errors for subsequent corrective action. However, our results reveal that while ChatGPT can produce some valid topical subject headings that conform to the LCSH vocabularies, it often fails to produce complex subject strings that adhere to intricate subdivision rules.

When addressing the issue of validity of the LCSH generated by ChatGPT, catalogers have the following four possible solutions at their disposal:

1. Manually search for alternative or broader subject heading terms through additional research into LCSH controlled vocabularies and authority records;

2. Use existing software for rectification. Reconciliation tools can assess ChatGPT-generated subject strings, distinguishing correct LCSH from invalid ones. This method is particularly effective for addressing the five aforementioned categories of validity issues. Again, ILS can be used for a routine check for errors in MARC records that have been enriched with these subjects. In addition, open-sourced reconciliation software such as *Library of Congress OpenRefine Reconciliation Service Endpoint,* [4] are also readily available for use.

3. Employ faceted vocabularies that do not require intricate rule interpretation and application for subject assignment. A faceted vocabulary term represents only a single concept. While some faceted subject strings include subdivisions, e.g., *Sports--Political aspects* from FAST (Faceted Application of Subject Terminology), all components are in the same subject category. To express multiple concepts, multiple faceted vocabulary terms are to be assigned. Therefore, there is no need for rule interpretation and application. This will certainly make subject assignment much more straightforward and yield better results.

4. Propose new subject authority records for LLM-generated subject headings that would provide a more precise or nuanced description than those available in the LCSH controlled vocabulary (please see category D in our results of validity evaluation) through the Subject Authority Cooperative (SACO) of the Program for Cooperative Cataloging. While proposals require adherence to the SHM rules and guidelines, individuals can seek support from a SACO funnel[5] with relevant expertise.

Although the LLM we used in the present study was ChatGPT, which is not an open-source LLM, so its exact training data are not known, our results provide strong evidence that ChatGPT has been trained on LCSH vocabularies scraped from the Library of Congress website and MARC records from websites of libraries and publishers around the world. While, as a closed source large language model, ChatGPT cannot be retrained by individuals to improve the accuracy of LCSH assignment, the recent emergence of open-source LLMs, such as Llama and Mistral, may present an additional opportunity to train custom models using the LCSH controlled vocabularies or a large number of quality bibliographic data from diverse disciplines. Bibliographic records with the Program for Cooperative Cataloging (PCC) authentication are considered quality data.[23] PCC contributors are required to go through a rigorous training and review process before being approved to create PCC authenticated records on their own. In addition, PCC records cover a broad range of disciplines and topics, even though the content of resources represented by these records tends to be more academic. Using these records to train an open-source LLM might significantly increase the quality of subject assignment.

---

[4] See https://github.com/cmharlow/lc-reconcile

[5] See https://www.loc.gov/aba/pcc/saco/funnels.html

Our results demonstrate that specificity and exhaustivity of LCSH assigned by ChatGPT vary considerably across records – to the effect that only about half of ChatGPT's output satisfactorily meets both criteria. It is worth noting that, unlike MARC coding accuracy and LCSH validity that can be checked against the rigid rules and controlled vocabularies, evaluation of specificity and exhaustivity is highly subjective. Subject analysis is not an exact science, but rather a blend of activities prone to subjectivity, such as comprehension, interpretation, and even translation. It is not uncommon for two catalogers to assign different subjects headings to the same resource due to their unique perspectives. Indeed, as previously demonstrated by Tonta,[24] the subject headings assigned by catalogers from the Library of Congress and the British Library to a set of identical books were mostly inconsistent with each other. Last but not least, specificity and exhaustivity of subject headings are also affected by external factors, such as institutional policies, practices, and time constraints.

**Conclusion**

In this study, we explored the use of large language model (LLM), specifically ChatGPT, to generate Library of Congress Subject Headings (LCSH) for electronic theses and dissertations (ETDs). The results revealed that, while ChatGPT can access an internalized corpus of the LCSH and MARC 21 records, the model struggles with validity, specificity, and exhaustivity in the generated subject headings. Furthermore, while validity issues can to some extend be mitigated using complementary tools, the model still falters in terms of specificity and exhaustivity, particularly with LCSH strings that involve subdivisions, thus failing to produce accurate results about half of the time. These results suggest that, in order to ensure accuracy and reliability of the cataloging process, the involvement of human catalogers remains an essential prerequisite.

This having been said, the use of LLMs like ChatGPT in cataloging ETDs or similar materials whose titles and abstracts are highly indicative of their topics can still play an important supportive role. Today, many academic libraries are overwhelmed with an influx of new electronic resources, including ETDs, that by far exceed human catalogers' capacity to process in a timely manner. Our study suggests LLMs could be used as an economical tool for the immediate generation of the LCSH given the cost associated with using Microsoft's OpenAI API in the present study amounted to approximately USD $0.25 and the total time used for processing the 30 ETDs was roughly 3 minutes. As an initial filtering and suggestion tool, ChatGPT allows experienced catalogers to focus on enhancing the accuracy and completeness of the catalog records. This complementary approach can potentially boost productivity by

leveraging LLMs for basic tasks while preserving the critical, irreplaceable expertise of human catalogers. Furthermore, for junior catalogers in particular, refining an existing, albeit imperfect, LCSH suggestion from ChatGPT is less daunting than constructing new subject headings from scratch. Therefore, using LLMs has potential to reduce cataloging time needed to assign LCSH subject terms for ETDs. In addition, for LLM-generated subject headings that are not valid LCSH, but may still be sufficiently descriptive enough for describing, catalogers may even consider inserting these terms directly in the 653 (uncontrolled index term) field in a MARC record to improve discoverability of the work in a library catalog.

The present study has several limitations. First, our approach was not a rigidly designed scientific experiment with a large, randomized sample of material to be cataloged. Accordingly, a larger study is merited to examine the consistency of ChatGPT output for assigning LCSH to a larger sample size. Second, our results may only be narrowly applicable to the specific type of resources — namely, to theses and dissertations that often include abstracts and titles that are highly indicative of their topics. However, this is not the case with many other types of resources in an academic library collection. Finally, considering that, in the present study, we focused exclusively on the generation of the LCSH, future experiments could explore subject assignment in other frameworks, such as FAST (Faceted Application of Subject Terminology), as well as in other types of works (books, audio-visual materials, etc.), which will provide further insights into the applicability of LLMs in library cataloging.

**Acknowledgements**

We gratefully acknowledge the valuable assistance of Ms. Lolita Kwok (Head of Resource Discovery) and Mr. Anthony Wong (Library Assistant) from the Hong Kong Baptist University Library.